\documentclass[Afour,sageh,times,doublespace]{sagej}

\usepackage{moreverb,url}

\usepackage[bookmarksopen,bookmarksnumbered,citecolor=black,urlcolor=black]{hyperref}

\newcommand\BibTeX{{\rmfamily B\kern-.05em \textsc{i\kern-.025em b}\kern-.08em
T\kern-.1667em\lower.7ex\hbox{E}\kern-.125emX}}

\def\volumeyear{2023}

\usepackage{setspace}
\usepackage{algorithm}
\usepackage{algpseudocode}
\usepackage{makecell}

\begin{document}

\runninghead{Cuan et al.}

\title{Interactive Multi-Robot Flocking with Gesture Responsiveness and Musical Accompaniment}

\author{Catie Cuan\affilnum{1, 2}, Kyle Jeffrey\affilnum{3}, Kim Kleiven\affilnum{2}, Adrian Li\affilnum{4}, Emre Fisher\affilnum{3}, Matt Harrison\affilnum{5}, Benjie Holson\affilnum{2}, Allison Okamura\affilnum{1}, and Matt Bennice\affilnum{5}}

\affiliation{\affilnum{1}Stanford University, United States\\
\affilnum{2}Independent Engineer (Work completed while a full-time employee Everyday Robots, United States)\\
\affilnum{3}Virtusa Corporation, United States (Work completed while on contract at Everyday Robots, United States)\\
\affilnum{4}Wayve, United States (Work completed while at Everyday Robots, United States)\\
\affilnum{5}Google Inc., United States\\}

\corrauth{Catie Cuan, Stanford University
CHARM Laboratory,
440 Escondido Mall,
Stanford, California
94305, United States}

\email{ccuan@stanford.edu}

\begin{abstract}
For decades, robotics researchers have pursued various tasks for multi-robot systems, from cooperative manipulation to search and rescue. These tasks are multi-robot extensions of classical robotic tasks and often optimized on dimensions such as speed or efficiency. As robots transition from commercial and research settings into everyday environments, social task aims such as engagement or entertainment become increasingly relevant. This work presents a compelling multi-robot task, in which the main aim is to enthrall and interest. In this task, the goal is for a human to be drawn to move alongside and participate in a dynamic, expressive robot flock. Towards this aim, the research team created algorithms for robot movements and engaging interaction modes such as gestures and sound. The contributions are as follows: (1) a novel group navigation algorithm involving human and robot agents, (2) a gesture responsive algorithm for real-time, human-robot flocking interaction, (3) a weight mode characterization system for modifying flocking behavior, and (4) a method of encoding a choreographer's preferences inside a dynamic, adaptive, learned system. An experiment was performed to understand individual human behavior while interacting with the flock under three conditions: weight modes selected by a human choreographer, a learned model, or subset list. Results from the experiment showed that the perception of the experience was not influenced by the weight mode selection. This work elucidates how differing task aims such as engagement manifest in multi-robot system design and execution, and broadens the domain of multi-robot tasks.
\end{abstract}



\maketitle

\section{Introduction}

\begin{figure}[h!]
\centering
\includegraphics[width=\linewidth]{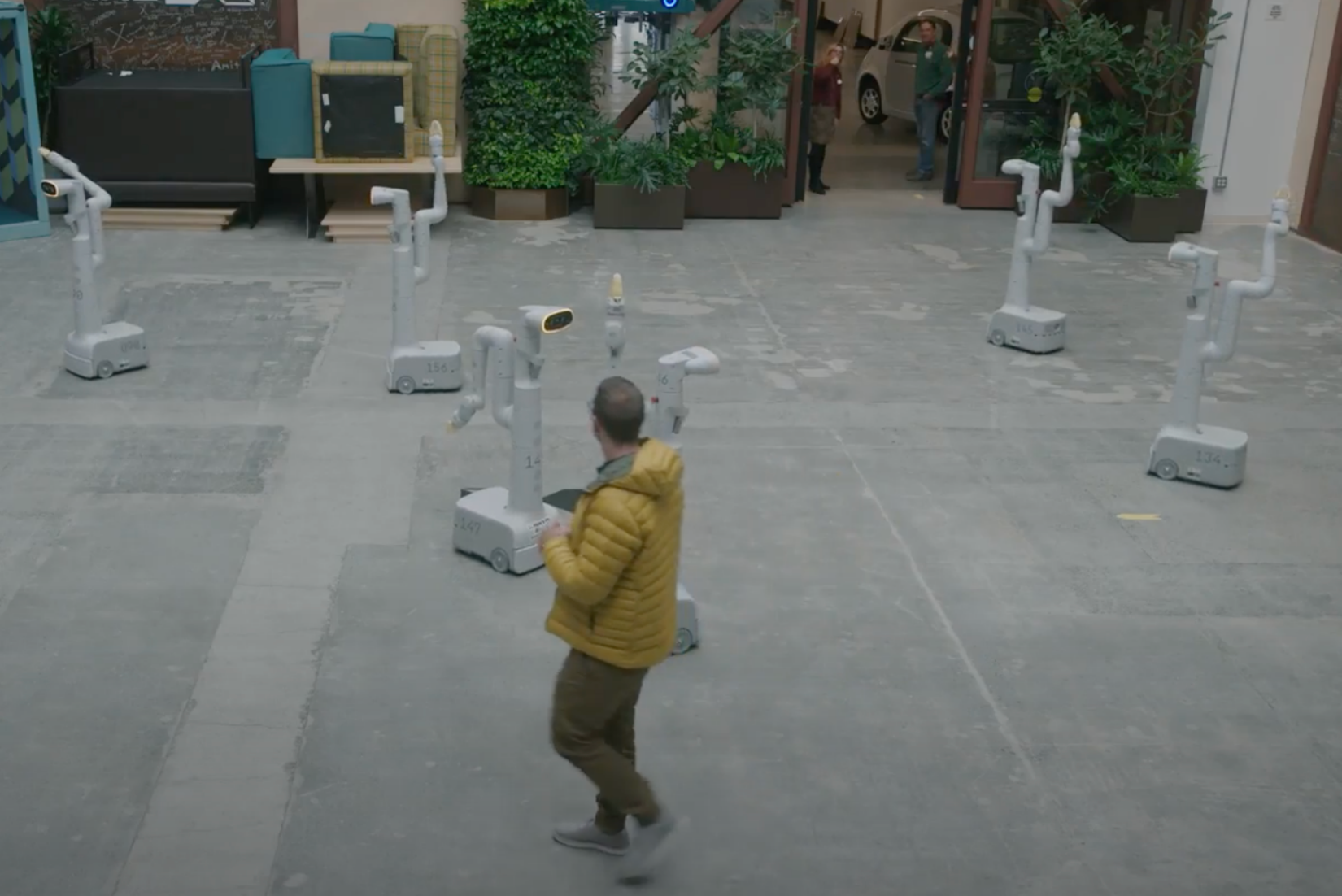}
\caption{Autonomously flocking robots move nearby a participating human in this interactive experience.}
\label{glamournew}
\end{figure}

In the past, robots have been limited to commercial and research spaces, where trained roboticists and programmers were the primary robot users and interactants. As robots have begun to exit these spaces and enter everyday environments like homes, offices, restaurants, and hospitals, new human-robot experiences and modes of interaction have proliferated. Robots that have not been designed for explicit social purposes \textit{become} social due to their placement in a varying, everyday human environment. This phenomenon then begs the question -- what kinds of novel experiences can we create that leverage a robot's features as both a utilitarian tool and an evocative, social agent?

The personal computing revolution demonstrated a similar trend. As computers became smaller and more accessible, they were leveraged for a vast number of activities, from communication to artistic expression, gaming to record keeping. As these activities expanded, the form factors, auxiliary tools, interfaces, and cultural significance of the personal computer shifted as well. Diverse, new academic subdomains proliferated, such as Human-Computer Interaction and Computer Graphics. Several such academic disciplines and sub-disciplines exist in robotics: Human-Robot Interaction, Social Navigation, and Human-Centered Robot Design are a sampling. This work examines how human-robot interaction modes like gestures and movement can be designed and built for an everyday environment and engaging social setting.

Multi-robot cooperation is an approach to tackle various tasks. One paper categorized multi-robot systems (MRS) by application domain: Surveillance and Search and Rescue, Foraging and Flocking, Formation and Exploration, Cooperative Manipulation, Team Heterogeneity, and Adversarial Environment \citep{darmanin2017review}. Multi-robot systems are typically beneficial in cases where the task is too complex for a single robot to achieve or where destruction risk is high and thus simplistic robots with redundant capabilities may be preferred. In this work, the task of creating a compelling interaction was amplified by increasing the number of robots. As one aim was for the system to be continuous and improvisational, supervised learning was incorporated in the gesture detection and the base motion. 
 
Robotics research is motivated by many real--world applications, such as agriculture, search and rescue, manufacturing, surgery, and construction. One categorization scheme for these varying motivations is to group them into two overlapping primary categories: improving human quality of life and generating economic value. Within those criteria, art and performance is an impactful pursuit. The size of the art and culture market in the United States in 2022 was roughly 400 billion dollars \citep{SizeArtMarket}. A 2017 Survey of Public Participation in the Arts (SPPA) found that in the last year, 54\% of U.S. adults created or performed art, 54\% attended artistic, creative, or cultural activities, and 10\% formally learned an art \citep{ArtsParticipation}. According to a survey from Ipsos on behalf of Americans for the Arts, 68\% of Americans agree that the arts have a positive effect on their health and well-being \citep{ForTheArts}. Given this economic and social footprint, using robots for the generation of and participation in art/performance, is a consequential research aim. 

Thus, the primary motivating factors of this work are summarized as: (1) Examining and increasing compelling interactions between humans and robots as they exit research and commercial spaces and enter everyday, social ones, (2) Building a combined learned and semi-controlled multi-robot software system, and (3) Creating innovative, beautiful, and influential artistic work that magnifies human and robot expression. The team included an artist/choreographer, software engineers, and a music composer. 

Given these motivations, a team consisting of an artist/choreographer, software engineers, and a music composer wrote algorithms and software for a group of robots to interactively flock with a human. The team foregrounded the overall engagement of the experience by supplementing the robot's movement with Music Mode, a software package that transformed the robot's movement into music \citep{cuan2023music}. The final experience runs on several robots in the real world and has been demonstrated dozens of times for audiences of varying sizes (in a non-experimental setting). The contributions of this work are as follows: (1) a novel group navigation algorithm involving human and robot agents, (2) a gesture responsive algorithm for real-time, human-robot flocking interaction, (3) a weight mode selection system for modifying flocking behavior, and (4) a method of encoding a choreographer's preferences inside a dynamic, adaptive, learned system. An experiment was also designed to study participants' overall impression of the flock when flocking behavior was determined by a choreographer, a learned model, or an unvaried condition.
\section{Prior Work}

\subsection{Multi-Robot System Design}
Parker described ``distributed intelligence'' and its application to systems of agents that can achieve tasks as a team. He wrote that ``some applications can be better solved using a distributed solution approach –- especially those tasks that are inherently distributed in space, time, or functionality.'' \cite{parker2007distributed} further wrote that robotic tasks with parallel problems, redundant agents, or the inability to create a single monolithic entity to solve it, may be better suited to multi-robot systems. Survey articles detail multi-robot systems in manipulation \citep{feng2020overview} and path planning \citep{madridano2021trajectory}, as well as general prior research \citep{gautam2012review, parker2016multiple} and networking strategies \citep{jawhar2018networking}. These surveys share several common assertions about ongoing challenges: optimization under shared constraints and developing learning or decision-making algorithms with high flexibility. Heterogeneous multi-agent systems are surveyed in \citep{rizk2019cooperative}, where the authors divided prior work along the categories of task decomposition, coalition formation, task allocation, perception, and multi-agent planning and control.

In this project, robots are cooperative and share the same task goal, as contrasted with other multi-robot systems papers that explore task identification, decomposition, and allocation. This work differs from many other multi-robot system tasks in that the domain-specific success or goal is to elicit a positive emotional response and comfortable interaction in humans. The near-term goals for each robot are to drive to different locations based on a series of calculations, but the broad goal of all the robots in the combined task is to be enjoyable to the humans participating in the interaction and observing nearby.

Prior works have described original software languages written explicitly for swarm robot control; these include Buzz \citep{pinciroli2016swarm}, IAda \citep{duhaut1991including}, and Protoswarm \citep{bachrach2008protoswarm}. Researchers have also programmed swarm drone performances \citep{kim2016realization, kim2018aeroquake}. In this work, all software (classes, methods, etc.) was original and written in Python. Throughout this project, the number of robots deployed was typically 6-7, so the word swarm is replaced with flock, as other works define swarm as $N > 10$ robots.

\subsection{Human-Swarm Interaction and Gesture Responsiveness}
Human-Swarm Interaction (HSI) is a field related to Human-Robot Interaction (HRI), where a single or many humans are interacting with many robots simultaneously. Prior work has explored how increasing the number of interacting robots alters cognitive and physical human load. Researchers found that human psychophysical response was stronger for larger numbers of robots \citep{podevijn2016investigating}. This work showed that increased psychophysical response by testing different groups of robots. In an HSI survey, \cite{kolling2015human} posited that in a case where homogeneous robots are coordinating autonomously, human control (or interaction) with the swarm will be $O(1)$, the same as a human interacting with a single robot. In cases where the robots are acting independently, the authors wrote that the human load will be $O(N)$ over the $N$ robots.

Given the difficulty of mapping one human's interactions to $N$ robots' actions, several researchers have explored gesture as a method to control and interact with multiple robots \citep{canal2015gesture}. \citep{kim2020user} conducted a study to determine user preferences for commanding a swarm to perform different behaviors, and found that gesture was the most commonly used interaction modality. \citep{alonso2015gesture} extended their work on robotic displays to include a gesture-detecting sensor setup, such that a participant controlled the robots' position, movement, and color using gestures. \cite{couture2010selecting} explored similar ideas with gesture control in a multi-robot system. \cite{de2022gestures} implemented gesture patterns for teleoperation control of a heterogeneous multi-robot system. \cite{berger2021exploring} found that certain human gestures led to similar human interpretations when the gestures led multiple robots to act. 

In this work, the expressive movement effectively \textit{is} the robot task; the robots' ability to be beautiful and compelling through their motion is one of the primary task aims. Others have also explored how individual robots and swarms of robots can convey emotion, images, and expression. \cite{simmons2017keep} defined expressive motion as movements that are not specifically task-oriented, but ``help convey an agent's attitude towards its task or its environment, such as emotional state, confidence, engagement, etc.'' \cite{santos2021motions} conveyed fundamental emotions with an expressive robot swarm. \cite{alonso2012image} built a novel display from a group of mobile robots to showcase entertaining images and animations.

Other researchers have studied how robot motion is perceived without the lens of expressive motion. \cite{dietz2017human} measured the effects of speed, smoothness, and synchronization on valence, arousal, and dominance. \cite{kim2021generating} studied swarm motion legibility, and found that behavior-based swarm motion was the most legible to participants, whereas trajectory-based legible motion was most effective for glanceability. Research with one robot found that predictability and legibility are different and often contradictory \citep{dragan2013legibility}.

Researchers have also aimed to build multi-robot systems that generated a connection between humans and robots. \cite{villani2020humans} coupled gestures with heart-rate detection to create a natural affect system for human-swarm interaction. \cite{st2019engaging} established a bond between multiple robots and an operator through full-body expressive motion. In other work to build a connection between humans while interacting with multiple robots, researchers made an interface for multiple humans to interact simultaneously while interacting with multiple robots \citep{patel2021direct, patel2020improving}.

\subsection{Choreographic and Theatrical Methods in Robotics}
Choreographers design movement for humans, and roboticists design movement for robots. Given these shared aims of generating movement, roboticists and choreographers have collaborated on research and artistic projects. \cite{apostolus1990robot} choreographed some of the earliest dances for robots in the 1980s. She motivated and defined ``robot choreography'' as ``Robotic movement in this [traditional] sense is not an abstraction from reality but rather than assistant to everyday action... Robot choreography poses a different approach: the development of deliberately choreographed movement in a nonwork environment and unrelated to utilitarian tasks.'' This work defines robot choreography differently -- as a practice where aesthetically choreographed motion (with attention towards smoothness, continuity, and narrative) is paired with the achievement of everyday tasks (like navigation) such that the robot's overall presentation is contextually and socially harmonious.

\cite{laviers2018choreographic} detailed several theoretical and practical applications for choreographic robot programming and design. An interdisciplinary research team of choreographers and roboticists designed an experimental testbed for human-robot interaction within dance performances and artistic installations, and widely performed this work \citep{cuan2018curtain, cuan2018time}. Further work from that team studied how moving with robots in an artistic installation affected how the robots were perceived \citep{cuan2019time}. \cite{hoffman2014designing} advocated on behalf of movement--centric design when making expressive social robots.

Robots have appeared extensively in live dance and theater performances and installations in the last couple decades, a small collection of noteworthy and similar works are listed here. \cite{breazeal2003interactive} created an interactive robot theater installation at SIGGRAPH in 2002 in order to engage conferences attendees. \cite{lin2009realization} used many robot form factors to realize a fully-robot theater. A contemporary choreographer used KUKA robots to replicate a human dance performance while grasping black flags \citep{forsythe2014black}. In \citep{ladenheim2020live}, an artist worked with robotists to create a wearable robot for use in live dance performance. 

In a similar vein, roboticists have collaborated with theater artists and drawn inspiration from theatrical improvisation methods \citep{knight2011eight, rond2019improv, sirkin2014using}. These prior improvisation games and strategies allowed researchers to quickly generate and refine new expressive movement profiles for robots. This project includes a similar interdisciplinary team and arts and roboticists, where improvisation is incorporated through a learned model, rather than a game or strategy. 

\subsection{Robots and Music}
Robots have played and acted as musical instruments. In prior work, researchers and musicians built original robots as musical instruments \citep{singer2004lemur, singer2005large, singer2003lemur, murphy2014expressive, kapur2011machine}. Robots have been explicitly designed in order to play instruments like the marimba \citep{hoffman2010shimon}, keyboard \citep{kim2011enabling}, and Theramin \citep{mizumoto2009thereminist}. In prior improvisational work, software enabled the robot to improvise with a human musician \citep{weinberg2009interactive, hoffman2010shimon}.

Researchers investigated how different sounds produced by robots affect human perception of the robot. \cite{hoffman2013effects} determined that a robot speaker playing music with synchronized movement led to more positive human-like traits than playing the music without matching movement. Studies of servo motors incorporated in robots found that humans may dislike the sound and robot producing it \citep{moore2017making, moore2019unintended}. Certain sounds may be useful for determining a robot's location \citep{cha2018effects} or its next action \citep{moore2020sound}. \cite{robinson2022designing} described sonic design principles towards achieving different aims; the SONAO project has similar goals \citep{bresin2021robust}. Here, we deployed movement-driven music software inside a group of robots, in order to create an engaging, interactive music and choreography experience.  

\section{System Design}

The robots in use are prototype robots designed and manufactured by Everyday Robots in Mountain View, California. The robots have a 7-degree-of-freedom arm, a pan-tilt head, a two-finger gripper, and a non-holonomic, mobile base. Each robot has a map of the interior environment and uses LiDAR to localize within the space. The robots have speed and force limits, as well as compliance and obstacle avoidance algorithms to make them human-friendly. They emit sound via a single speaker in the front of the robot's main torso column.

\begin{figure}[t]
\centering
\includegraphics[width=\columnwidth]{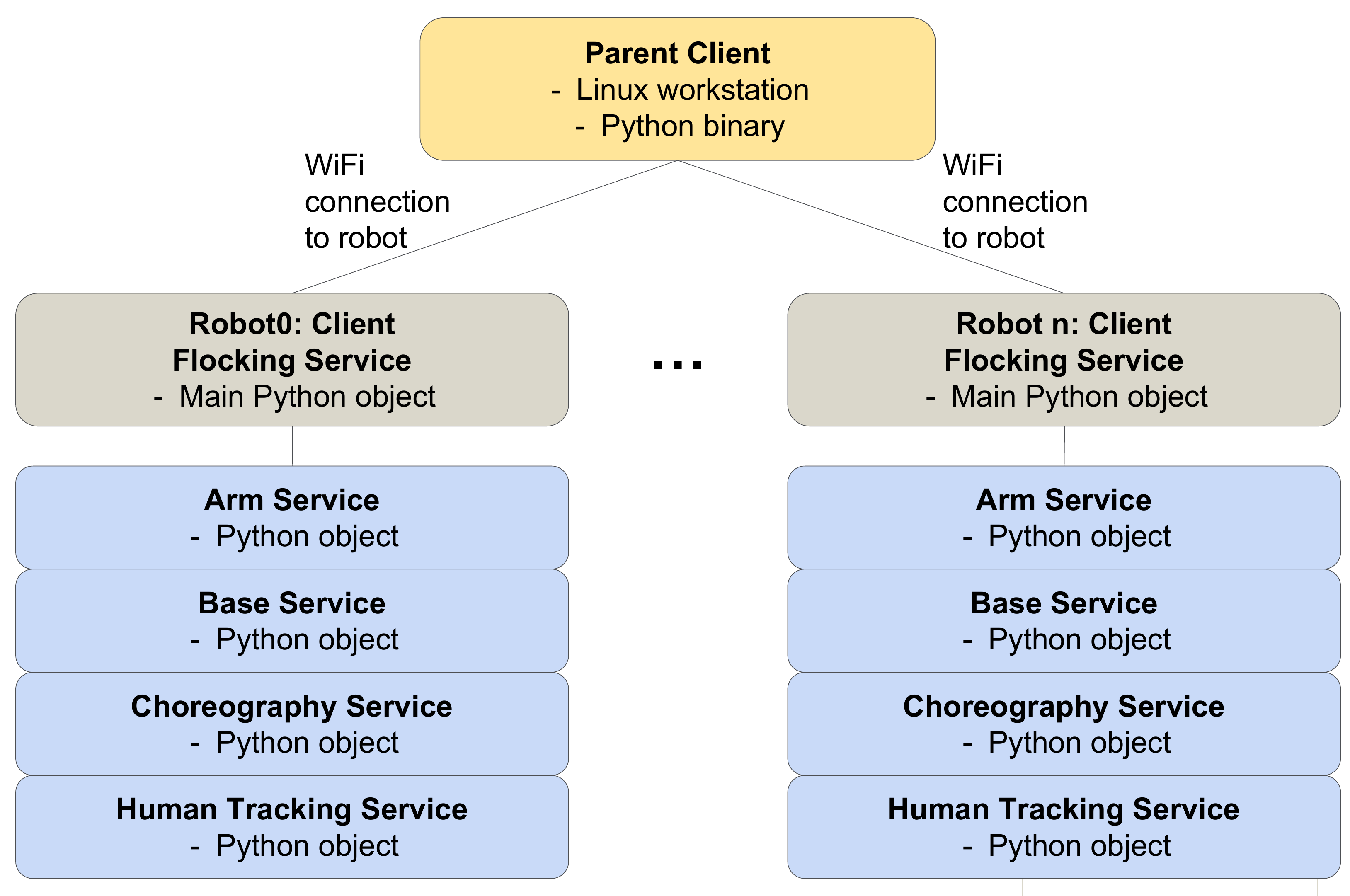}
\caption{The overall flocking system. Individual robots connect to the parent client running on a workstation. This parent client creates individual flocking service objects for each robot, which includes 4 subservices to manage different behaviors.}
\label{parentchild}
\end{figure}

Within the flock, the robots are modeled as homogeneous, cooperative agents. The humans in the same boundary region as the robots are incorporated into the flock, thus the overall system is comprised of heterogeneous agents: humans and robots. When the full system is running, each agent continuously publishes its navigation and sensor reading information to a parent client. The parent client exists as a Python binary, runs on a workstation, and connects to each robot client in the flock via ssh port forwarding. A flock consists of 1-10 robots, usually 6 or 7 given the physical space constraints in this project. The parent client combines the individual robots' information and shares this with each individual robot client. At the start of a flocking episode, the parent client creates an individual flocking service Python object associated with each robot and stores these as values in a dictionary with corresponding robot key names. The individual flocking service object creates four subservices that are associated with the first three of the four main robot subsystems, see Figure \ref{parentchild}.

\begin{figure}[t]
\centering
\includegraphics[width=\columnwidth]{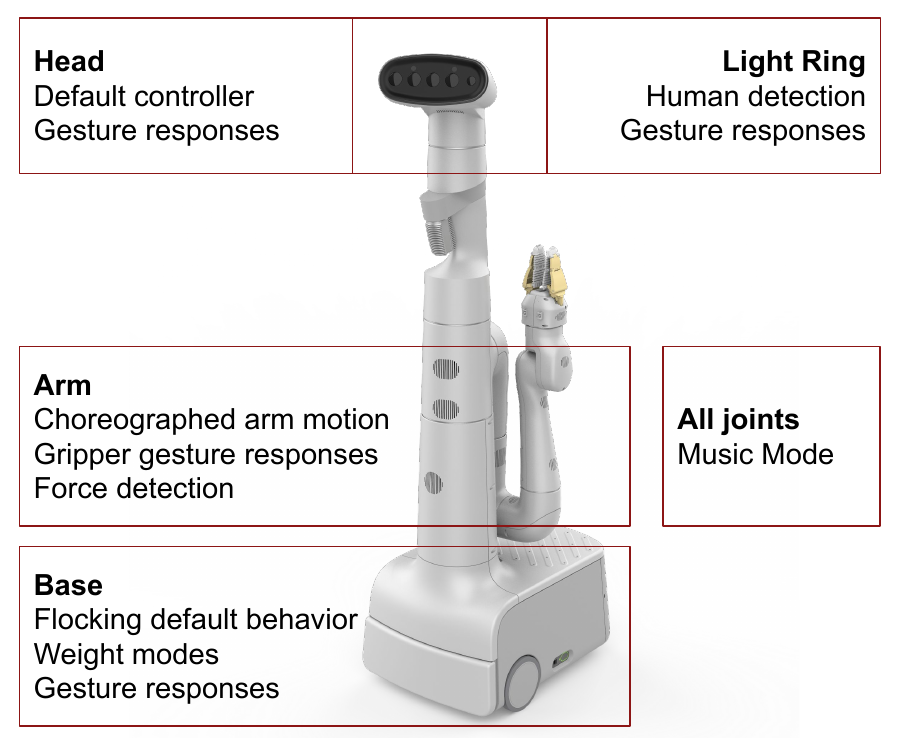}
\caption{The system components for an individual robot. The four main components are the Base, Arm, Head, and Music Mode.}
\label{individualcomponents}
\end{figure}

Each individual robot's behavior is divided into four main subsystems: 1) Head, 2) Arm, 3) Base, and 4) Music Mode, shown in Figure \ref{individualcomponents}. Each subsystem has a separate control scheme and all of them run in parallel. The fourth subsystem, Music Mode, is a separate optional software application that runs on the robot and transforms the robot's movement (joint encoder readings) into sound. Further information about this music-generation application can be found in a research paper from the team \citep{cuan2023music}. While the novel software and research described in this flocking paper does not expand upon or interact with the Music Mode software, Music Mode is included here as a subsystem because it is critical to the overall experience generated by the robot flock. The Music Mode software runs for the duration of the robot flocking and all the sound is generated in real time. As each of the robot's joints and the robot's base moves, corresponding composed sounds are generated and play through the robot's frontal speaker. The volume of each individual robot can be set and changed at any time.

In order to describe the overall behavior of the flock, the following terms will be used throughout the paper and are explicitly defined for reference below:
\begin{itemize}
    \item \textbf{Subsystem}: A general aspect of the individual robot's behavior. There are four subsystems: Head, Arm, Base, and Music Mode.
    \item \textbf{Individual flocking service}: A Python object associated with each robot in flock and instantiated by the parent client. The flocking service houses and references the subservices.
    \item \textbf{Subservice}: Four classes associated with the individual flocking service that track different aspects of the individual and group robot behavior. The four subservices are Arm Service, Base Service, Choreography Service, and Human Tracking Service.
    \item \textbf{Service step}: A single iteration of the flocking algorithm, which iterates each subservice in the individual robot flocking service.
    \item \textbf{Weight mode}: A set of scalars associated with each linear term in the base motion linear equation. Different weight modes result in different flocking behaviors. There are seven weight modes, to be discussed further in the Weight Modes and Supervised Learning Process subsection.
    \item \textbf{Boundary region}: A mapped region where the robots are flocking, comprised of x and y minima and maxima of the base.
\end{itemize}

The full robot flock runs at 20Hz. At every service step, each subservice is called and all state variables about robot positions, robot sensor values, detected human positions, and detected gestures are updated. Each subservice has a default behavior and this behavior can be interrupted and superseded by a new behavior in response to a detected human gesture. The flocking action area is constrained to a specific large, mapped area and this boundary region is incorporated into flocking calculations. This boundary region and all positions are calculated in a shared ``map'' frame.

\subsection{Head Controller}

A head controller in the parent client manages the movement of every robot head within the flock. This controller is updated once per service step and is a function of whether or not there are humans in the scene and whether or not they are gesturing. If there is a single person in the boundary region and they are gesturing, all robots will look at that person. If there are many humans in the scene who are all gesturing, the human tracking service will return a single human to look at per service step. If there are multiple people in the scene and one of them is gesturing, all robots will look at that person. If there are multiple people in the scene and none of them are gesturing, each robot will look at the person closest to it. If there are not people in the scene, the robots will look at the center of the region. Given this logic, gesturing humans are prioritized by the head controller, then present humans, and as a default behavior, the robots look towards the center of the boundary region.
\begin{figure}
\centering
\includegraphics[width=\linewidth]{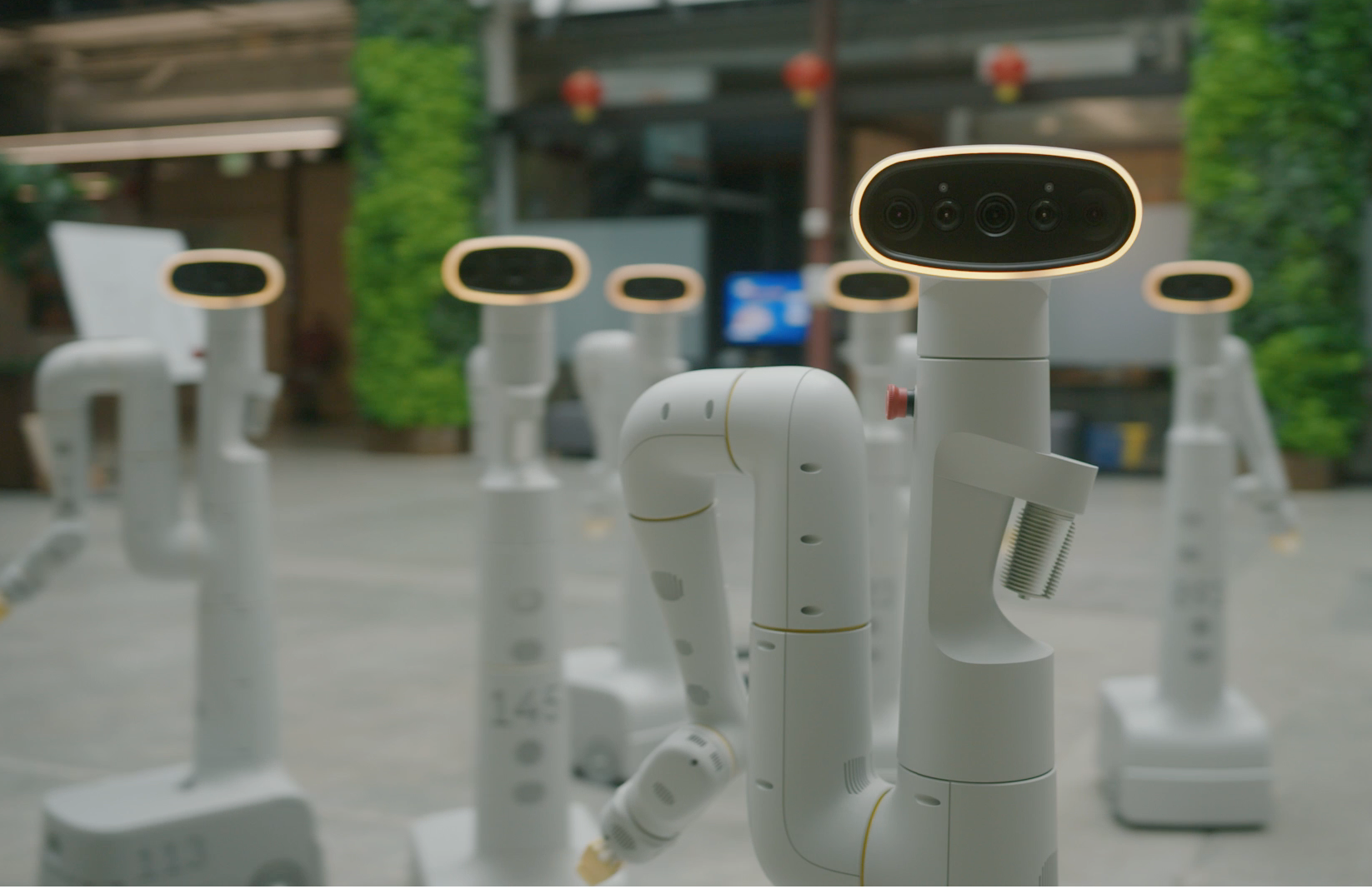}
\caption{The robots' light rings turn yellow when a human is detected in the scene. The head controller directs the robot to look at the closest human in the boundary area if one is detected.}
\label{lookingatyou}
\end{figure}

\subsection{Human Tracking Service}
The robots detect humans by using cameras in the head and LiDAR mounted mid-body. LiDAR samples are processed with TF3D Instance Segmentation \cite{boxfreenet} to find the 3D point cloud of each person, from which a bounding box is derived. A PersonLab model \cite{papandreou2018personlab} extracts 2D instance masks for each person in the RGB camera frame, and 2D positions of each body keypoint. The depth image is used to derive 3D point clouds and bounding boxes for each person, and 3D points for each person's keypoint (left shoulder, right shoulder, left wrist, right wrist, etc.). The bounding boxes from both camera and LiDAR detection pipelines are merged by a 3D tracker to estimate the dynamic state of each person. The body parts from both detection pipelines are not tracked in the same way, but every time its dynamic state is observed, it is spatially transformed so it stays attached to the person. At the output of this person detection-tracking subsystem, each detected human has an associated bounding box and relevant pose attributes, such as the location of their arms, legs, torso, and head in Cartesian space. While a human may be detected, not all humans will have the same associated body parts, such as both arms and both legs, as some parts may be occluded or outside the field of view of the sensors. Depending on the collected images, only one hand may be detected, for example. If the detected human is in the boundary region, the detected humans and their relevant attributes are stored in a dictionary by each individual flocking service. The humans are deduped such that if many robots detect the same human, only one human is added to each of the robots' individual dictionaries. 

The positions of the humans are incorporated into the base movement calculations, and in doing so, the humans in the boundary region effectively become part of the robot flock.

The robot heads include a track of programmable LED lights, referred to as a light ring. In this project, the lights are typically set to be all the same color at once. When the robots are flocking without any humans in the boundary region, the light rings are light blue. When humans are detected in the boundary region, the lights turn yellow, see Figure \ref{lookingatyou}.

\subsection{Base Service}

\subsubsection{Boids Algorithm}
The robots' base motion is dictated by an enhanced version of an effective and well-known flocking algorithm, Boids Algorithm \cite{reynolds1987flocks}. In the canonical Boids Algorithm, three terms are calculated and summed to dictate a single boid's motion: Cohesion, Separation, and Alignment. The terms ``boid'' and ``robot'' are interchangeable in this project, and robot will be used rather than boid. While the precise mathematical and software implementation details vary according to the application, the general concept of these three terms is as follows:
\begin{itemize}
    \item \textbf{Cohesion}: The robots stay together or move towards each other. In the equations below, this is denoted by $\mathbf{c}_i$ for robot $i$.
    \item \textbf{Separation}: The robots stay separate and move away from each other. Denoted by $\mathbf{s}_i$ for robot $i$.
    \item \textbf{Alignment}: The robots navigate towards the average heading of nearby other agents. Denoted by $\mathbf{a}_i$ for robot $i$.
\end{itemize}

These three terms are typically multiplied by different gains ($\{k_c, k_s, k_a\}$, respectively) and then summed together as a single calculated boid value to update the position $\mathbf{x}_i=[x_i,y_i]^\top$ (or velocity $\mathbf{v}_i$) of each robot $i$. For example: 
\begin{align} \label{boidvalue}
\Delta\mathbf{x}_i &= k_c\mathbf{c}_i + k_s\mathbf{s}_i + k_a\mathbf{a}_i \\
 \label{newrobotpositionboid}
\mathbf{x}_i &\gets \mathbf{x}_i + \Delta\mathbf{x}_i
\end{align}

While each robot's motion is updated according to this rule, in this project both robots and detected humans (within a boundary region) are included in the boid calculations. The term ``agent'' is used to refer to both robots and detected humans, and the calculations below apply across a flock of $n$ agents.

At every iteration, each robot $i$ calculates its own Cohesion term $\mathbf{c}_i$ as follows:

\begin{algorithm}
\setstretch{1.2}
\caption{Cohesion term $\mathbf{c}_i$ for robot $i$}\label{alg:cohesion}
\begin{algorithmic}
\State $\bar{\mathbf{x}} = \frac{1}{n}\sum_{j=1}^{n} \mathbf{x}_j$ \Comment{(average position of all agents)}
\State $\hat{\mathbf{x}}_i = \bar{\mathbf{x}} - \mathbf{x}_{i}$
\State $\mathbf{c}_i \gets \hat{\mathbf{x}}_i / \|\hat{\mathbf{x}}_i\|$
\end{algorithmic}
\end{algorithm}

The Separation term was initialized to $\mathbf{s}_i = \mathbf{0}$ and the minimum separation distance $d_{min}$ was 1.5 meters. Each robot $i$ calculates its own Separation term $\mathbf{s}_i$ as follows:

\begin{algorithm}
\setstretch{1.1}
\caption{Separation term $\mathbf{s}_i$ for robot $i$}\label{alg:separation}
\begin{algorithmic}
\State $\mathbf{s}_i \gets \mathbf{0}$  
\For{agent $j \neq i$}
\State $\Delta\mathbf{x}_{ij} = \mathbf{x}_{i} - \mathbf{x}_{j}$
\If{$\|\Delta\mathbf{x}_{ij}\| < d_{min}$}  
    \State $\mathbf{s}_i \gets \mathbf{s}_i - \Delta\mathbf{x}_{ij}\cdot(\|\Delta\mathbf{x}_{ij}\| - d_{min})$
    \State $\mathbf{s}_i \gets \mathbf{s}_i / \|\mathbf{s}_i\|$
\EndIf
\EndFor
\end{algorithmic}
\end{algorithm}

Each robot $i$ calculates its own Alignment term $\mathbf{a}_i$ as follows:

\begin{algorithm}
\setstretch{1.4}
\caption{Alignment term $\mathbf{a}_i$ for robot $i$}\label{alg:alignment}
\begin{algorithmic}
\State $\bar{\mathbf{v}}_i = \frac{1}{n}\sum_{j\neq i}^{n} \mathbf{v}_j$ \Comment{(average velocity of other agents)}
\State $\mathbf{a}_i \gets \bar{\mathbf{v}}_i / \|\bar{\mathbf{v}}_i\|$
\end{algorithmic}
\end{algorithm}

\subsubsection{Algorithmic Enhancements}
In order to enhance the base movement and introduce other behaviors into the robot flock, additional linear terms were incorporated to compute the final boid value: Following, Circling, Linearity, and Bounds Aversion. These terms increased variation and further distinguished the flock from classical, naturalistic flocking behavior to engaging, choreographic flocking behavior. The general concept of these four terms is as follows:
\begin{itemize}
    \item \textbf{Following}: The robots follow the first detected human in the boundary region. Denoted by $\boldsymbol{\phi}_i$ for robot $i$.
    \item \textbf{Circling}: The robots move along a circular path. Denoted by $\boldsymbol{\pi}_i$ for robot $i$.
    \item \textbf{Linearity}: The robots move along parallel lines. Denoted by $\boldsymbol{\lambda}_i$ for robot $i$.
    \item \textbf{Bounds Aversion}: The robots avoid the boundary region edges. Denoted by $\boldsymbol{\beta}_i$ for robot $i$.
\end{itemize}

Similar to the canonical Cohesion, Separation, and Alignment terms, these four enhanced terms are multiplied by different gains ($\{k_\phi, k_\pi, k_\lambda, k_\beta\}$, respectively) and then summed together as a single calculated boid value to update the robot's position $\mathbf{x}_i$ or velocity $\mathbf{v}_i$: 
\begin{align} \label{newboidvalue}
\Delta\mathbf{x}_i &= k_c\mathbf{c}_i + k_s\mathbf{s}_i + k_a\mathbf{a}_i \\ &+ k_\phi\boldsymbol{\phi}_i + k_\pi\boldsymbol{\pi}_i + k_\lambda\boldsymbol{\lambda}_i + k_\beta\boldsymbol{\beta}_i
\end{align}

\noindent As before, each individual flocking service calculates these terms for its current robot, relative to the other robots and detected humans in the boundary region.

At every iteration, each robot $i$ calculates its own Following term $\boldsymbol{\phi}_i$ relative to the position $\mathbf{x}_{human}$ of the first detected human from the human tracking dictionary:

\begin{algorithm}
\setstretch{1.2}
\caption{Follow term $\boldsymbol{\phi}_i$ for robot $i$}\label{alg:follow}
\begin{algorithmic}
\State $\hat{\mathbf{x}}_i = \mathbf{x}_{human} - \mathbf{x}_{i}$
\State $\boldsymbol{\phi}_i \gets \hat{\mathbf{x}}_i / \|\hat{\mathbf{x}}_i\|$
\end{algorithmic}
\end{algorithm}

The Circling term $\boldsymbol{\pi}_i$ was calculated as follows, where $\mathbf{p}_{center}=[x_{center},y_{center}]^\top$ is a two-dimensional vector corresponding to the center of the room in a map frame, and the radius $r=0.9\cdot\min(l,w) / 2$ where $l$ and $w$ are the length and width of the boundary region respectively. The current timestamp is $t$ and the period $T_\pi$ is 50 seconds.

\begin{algorithm}
\setstretch{1.1}
\caption{Circling term $\boldsymbol{\pi}_i$ for robot $i$}\label{alg:circling}
\begin{algorithmic}
\State $\theta_i = 2\pi\cdot \left(\frac{t}{T_\pi}+\frac{i}{n}\right)$
\State $\hat{\mathbf{x}}_i = \left(\mathbf{p}_{center} +
\begin{bmatrix}
r\cos\theta_i \\
r\sin\theta_i
\end{bmatrix}
\right) - \mathbf{x}_{i}$  
\If{$\|\hat{\mathbf{x}}_i\| < 2$}
\State $\boldsymbol{\pi}_i \gets \hat{\mathbf{x}}_i$
\Else
\State $\boldsymbol{\pi}_i \gets 2\cdot\hat{\mathbf{x}}_i / \|\hat{\mathbf{x}}_i\|$
\EndIf
\end{algorithmic}
\end{algorithm}

The Linearity term $\boldsymbol{\lambda}_i$ was computed similarly to the Circling term, generating an ellipse-like path. The period $T_\lambda$ was 37 seconds. Each robot is assigned a ``robot lane'' (indexed by $k_i$) based on its position sorted along the $x$-axis.

\begin{algorithm}
\setstretch{1.2}
\caption{Linearity term $\boldsymbol{\lambda}_i$ for robot $i$}\label{alg:linearity}
\begin{algorithmic}
\State $\theta_i = 2\pi\cdot\frac{t}{T_\lambda}$  
\State $k_i = (\arg\mathrm{sort}\;{x_j})_i$ \Comment{(``robot lane'' index)}
\State $\hat{\mathbf{x}}_i =
\begin{bmatrix}
0.75\cos\theta_i + w\cdot k_i/n + 2\\
y_{center} + l/2\cdot\sin\theta_i
\end{bmatrix} - \mathbf{x}_{i}$
\If{$\|\hat{\mathbf{x}}_i\| < 2$}
\State $\boldsymbol{\lambda}_i \gets \hat{\mathbf{x}}_i$
\Else
\State $\boldsymbol{\lambda}_i \gets 2\cdot\hat{\mathbf{x}}_i / \|\hat{\mathbf{x}}_i\|$
\EndIf
\end{algorithmic}
\end{algorithm}

The Bounds Aversion term $\boldsymbol{\beta}_i$ was computed as follows, where $[x_{min}, x_{max}]$ and $[y_{min},y_{max}]$ are the ranges of the $x$- and $y$-dimensions respectively for the boundary region. The margin value $m$ varied depending on the size of the boundary region.

\begin{algorithm}
\caption{Bounds Aversion term $\boldsymbol{\beta}_i$ for robot $i$}\label{alg:bounds}
\begin{algorithmic}
\State $\boldsymbol{\beta}_i \gets
\begin{bmatrix}
\min(x_{max} - m, \max(x_{min} + m, x_i)) \\
\min(y_{max} - m, \max(y_{min} + m, y_i))
\end{bmatrix} - \mathbf{x}_{i}$
\end{algorithmic}
\end{algorithm}

\subsection{Gesture Responsiveness}

In addition to the default behavior for the arm, base, and head, the robots' behavior changes depending on the gestures that the humans in the scene perform. Thus, the detected humans in the scene are not only incorporated into all the term calculations for the base movement, they are perceived in order to trigger gesture responses. Incorporating the humans in these two manners (a third to be explained later in the paper) served the aim of increasing interactivity and engagement.  

\begin{figure*}[h!]
\centering
\includegraphics[width=\linewidth]{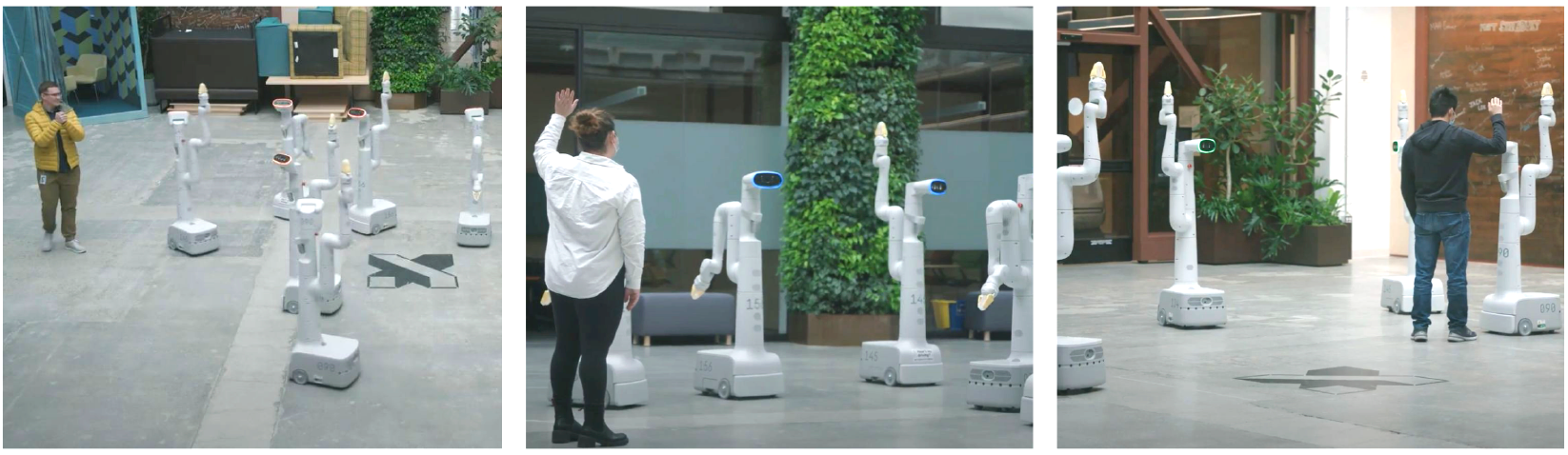}
\caption{Three gestures lead to different robot actions. Left: the ``Hands Together'' action, robots gaze upwards and the light rings turn orange. Center: the ``Left Hand Up'' action, robots open and close their grippers and the light rings turn dark blue. Right: the ``Right Hand Up'' action, robots turn in place and the light rings turn green.}
\label{gestures}
\end{figure*}

Three static gestures were designed in order to prompt different responses by the robots, see images in Figure \ref{gestures}. The three gestures and the relevant robot responses were as follows:
\begin{itemize}
    \item \textbf{Hands Together}: When a human's hands are together, (1) the light rings will turn orange and (2) the robots' heads will gaze up towards the ceiling and pan side to side. The hands are considered together if both of a human's hands are detected, and there is less than 0.08 meters of distance between the hands.
    \item \textbf{Right Hand Up}: When a human's right hand is up, (1) the light rings will turn green and (2) the robots will perform a 360 spin in place. If the robots are too close together to safely perform this spin, the light rings will pulse green repeatedly. A hand is considered up if it is higher than either of the human's shoulders.
    \item \textbf{Left Hand Up}: When a human's left hand is up, (1) the light rings will turn dark blue and (2) the robots' grippers will open and close.
\end{itemize}

The full gesture response behavior will complete on the robot's arm, head, or base, regardless of whether or not a new gesture is performed during the response time. In addition, the other default behaviors will continue in parallel to the gesture behaviors. For example, if a robot is engaged in a default flocking behavior (which includes a default behavior for the base, arm, and head) and a Left Hand Up is detected, the base and head will continue along their default flocking and head controller behaviors, respectively, as only the gripper reacts to the detected gesture.

In the programmed logic, Hands Together gestures are prioritized before Right Hand Up or Left Hand Up. Right Hand Up is prioritized if two hands were in the air. This was designated so that if a person has their hands together above their shoulders, this would be categorized as a Hands Together gesture rather than Right Hand Up and Left Hand Up. These gestures were selected because they were relatively simple to teach a new human interactant, the gesture could be performed in several ways (for example, a Right Hand Up could be directly over the interactant's shoulder, across their face, or diagonally to the side), and the gesture could be held for an extended period of time (versus a quick jump, which the robot's cameras may miss). 

The team experimented with different robot responses to the gestures and based the final gesture-to-robot-action on several criteria. The first was \textit{variation}: each gesture response should map to a different part of the main robot system (arm/gripper, head, and base). Robot responses were also selected to last different durations. The gripper opening and closing was a short duration at approximately 2 seconds, while the head gaze was medium at approximately 4 seconds, and the rotation in place was a long duration at approximately 12 seconds. 

The second criterion was \textit{comprehensiveness}. The team believed that an engaging robot response would include changes to the robot's light, motion, and sound.  The robot responses were coupled with light ring color changes in order to increase human recognition and certainty that the robot had received the correct gesture and would perform it. In addition, the robot responses were chosen to target a specific sound generation during Music Mode. For example, for Left Hand Up, the gripper opening and closing triggered a bell sound, this sound did not otherwise appear during the flocking because the gripper remained open. For Right Hand Up, the rotation in place continuously triggered the base motion sound, which loosely resembled a whirling. For Hands Together, the head gaze triggered a horn sound. 

The third criterion was \textit{harmony}: the robot response should have some relatedness to the overall behavior of the flocking and not be disjointed, annoying, or strange. The team selected this third criteria so the human interactant would be motivated to perform the gestures and would not be alienated by the robots' responses.

\subsection{Robot Arm Service}
The robots' arm movement consisted of four choices of choreographed arm moves. These sequence of joint configurations were used to generate a velocity--limited b-spline trajectory. The speed of the arm motion was 0.6 rad/second.

The Arm Service randomly selected between the four choices of choreographed arm trajectories. In this formulation, the gripper is treated separately from the arm motion. As none of the gesture responses required or interrupted the arm motion, the arm moves continued throughout the regular flocking behavior and gesture responses. The arm moves were choreographed based on two criteria: (1) maximizing the diversity and range of sounds generated by Music Mode and (2) not interfering with the front LiDAR sensor on the robot. As a result, many of the arm moves mostly exist behind the robot's front torso and have a wide range of motion.

\subsection{Weight Modes and Supervised Learning Process} \label{weightmodessection}

A weight mode is a set of scalars corresponding to the $\{k_c, k_s, k_a, k_\phi, k_\pi, k_\lambda, k_\beta\}$ gains that are multiplied by the different base motion terms in the individual robot's next position calculation. For example, a weight mode that generated exclusively circular motion would be $\{k_c=0, k_s=0, k_a=0, k_\phi=0, k_\pi=1, k_\lambda=0, k_\beta=0\}$, such that the computed next position value is:
\begin{equation} \label{circleboid}
\Delta\mathbf{x}_i = 1\cdot\boldsymbol{\pi}_i
\end{equation}

\begin{figure*}
\centering
\includegraphics[width=\linewidth]{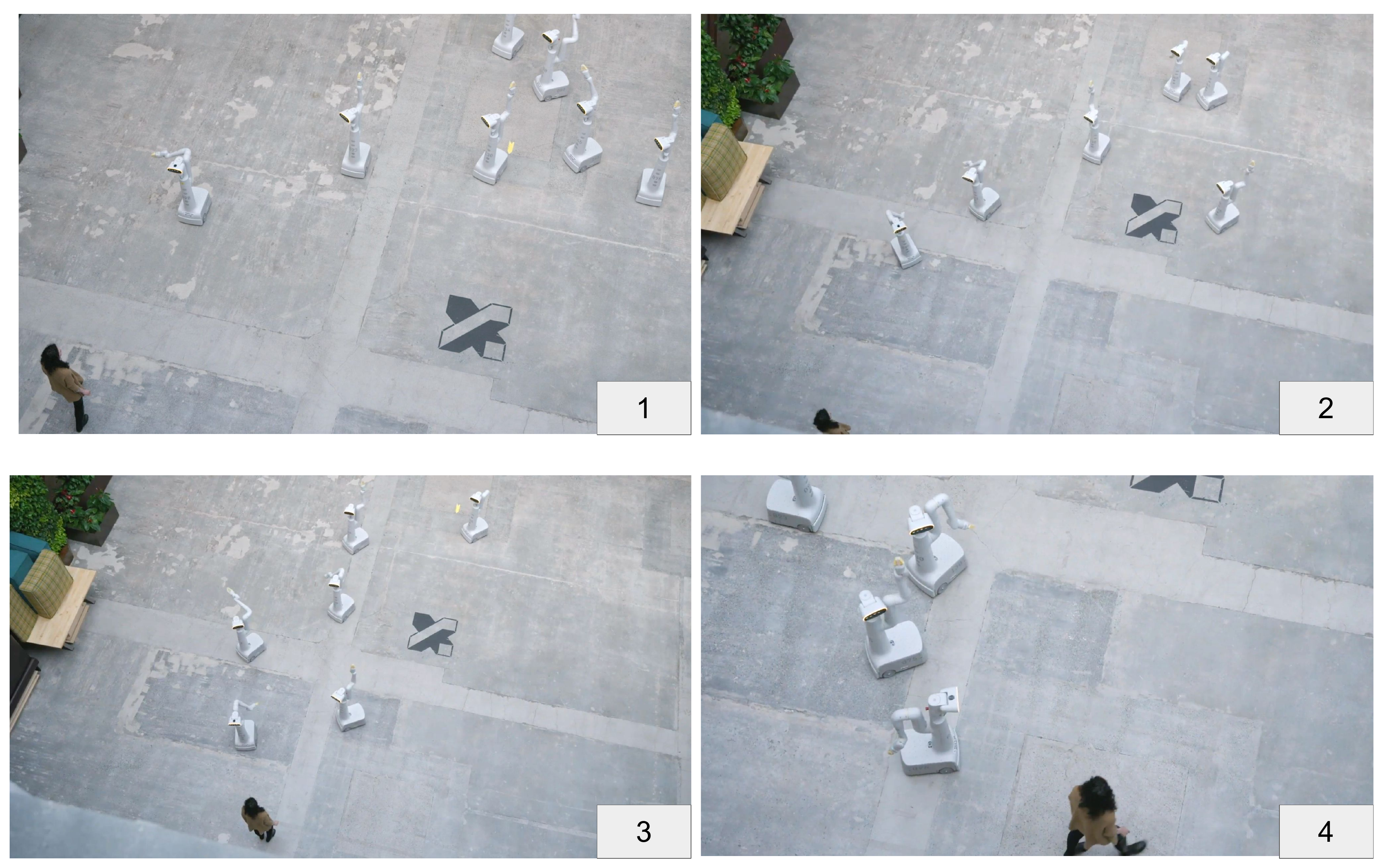}
\caption{A sequence of images from the ``Follow Human'' weight mode, in order of 1-4. The robots move towards the human at the lower left of the image while avoiding the edges of the boundary region.}
\label{followhuman}
\end{figure*}

The team experimented with several weight modes that aimed to create distinctive, compelling robot behaviors. Special attention was paid to the transitions into and out of each weight mode, such that the robot's behavior would be immediately and recognizably different. Each of the final seven weight modes included non-zero values for Bounds Aversion. The final seven weight modes were as follows, the positions of the robots while running each of these weight modes is illustrated in Figure \ref{weightmode_histogram}:
\begin{itemize}
    \item \textbf{Default}: All gains were assigned scalar weights. This led to behavior that appeared the most like conventional flocking, with some awareness and following of the humans in the scene.
    \item \textbf{Following}: The Following calculation was weighted most heavily. The robots were largely idle with no humans in the boundary region, and followed the humans closely when there were humans in boundary region, see Figure \ref{followhuman}.
    \item \textbf{Linear}: The Linearity calculation was weighted most heavily. The robots most frequently moved along lines, would avoid the boundaries, and avoid any humans in the boundary region.
    \item \textbf{Circling}: The Circling calculation was weighted most heavily. 
    \item \textbf{Cohesion}: The Cohesion calculation was weighted most heavily. If there were only robots in the boundary region, the robots would move towards each other and shift around while close. If there was a human in the boundary region, the Cohesion term factors this human in and the thus the robots would cohere both with each other and the included human.
    \item \textbf{Alignment}: The Alignment calculation was weighted most heavily. If there were no humans in the region or the human was still, the detected velocity was roughly zero so the robot movement was minimal. If there were humans in the region, the robots aligned with the humans' heading.
    \item \textbf{Separation}: The Separation calculation was weighted most heavily. The robots would move away from other robots and other humans in the boundary region.
\end{itemize}

Once these weight modes were created, the team ran the flocking behavior while an expert human choreographer selected between weight modes based on the robot and human behavior in the scene. The human choreographer aimed to make the robot behavior as engaging and varied as possible, especially relative to any humans that were present in the boundary region. For example, if the robots were moving in the Linear mode for 45 seconds and a gesturing human in the boundary area began to appear uninterested, the human choreographer might switch the weight mode to Following in order the reengage the gesturing human. As the goal was for the flocking behavior to be fully autonomous, the team decided to train a classifier that would select among weight modes based on the human and robot actions in the scene, just as the human choreographer did. Adding this feature would make the flocking behavior improvisational, reactive, and nondeterministic.

In this context, it is hard to define success of weight mode selection explicitly. There were three hand-engineered features that encoded geometric information about the world used as input to the model:
\begin{itemize}
    \item \textbf{Regional Density}: How the robots and humans in the boundary region are spread throughout the boundary region. For example, if all the robots and humans are crowded into a small corner of the full space, this value expresses that pattern.
    \item \textbf{Measure of Spread}: How the robots and humans in the boundary region are spread relative to each other.
    \item \textbf{Best Fit Line}: How well the robots and humans in the boundary region create a line.
\end{itemize}

The Regional Density $\boldsymbol{\rho}$ was computed as follows, where $\mathbf{x}_i=[x_i,y_i]^\top$ is the position of agent $i$, and $x_{max}$ and $y_{max}$ are the maximum $x$- and $y$-values of the boundary region respectively.
\begin{equation}
\bar{\mathbf{x}}  
\begin{bmatrix}
\bar{x} \\
\bar{y}
\end{bmatrix} = \frac{1}{n}\sum_i^n\mathbf{x}_i
\end{equation}
\begin{equation}
\label{regdensity}
\boldsymbol{\rho} 
\begin{bmatrix}
\bar{x}/x_{max} \\
\bar{y}/y_{max}
\end{bmatrix}
\end{equation}

The Measure of Spread $\sigma_i$ for robot $i$ was computed as follows:
\begin{equation} \label{measvec}
\sigma_i = \frac{1}{n}\left\|\sum_{j\neq i}^n(\mathbf{x}_i-\mathbf{x}_j)\right\|
\end{equation}

The Best Fit Line $\boldsymbol{\alpha}$ was computed as follows, where $n$ is the number of all humans and robots in the boundary area:
\begin{equation}
\mathbf{X} =
\begin{bmatrix}
\begin{matrix}
1 & x_1 & x_1^2 & x_1^3 \\
1 & x_2 & x_2^2 & x_2^3 \\
\end{matrix} \\
\vdots \\
\begin{matrix}
1 & x_n & x_n^2 & x_n^3 \\
\end{matrix}
\end{bmatrix}
\end{equation}
\begin{equation}
\mathbf{Y} =
\begin{bmatrix}
y_1 \\
y_2 \\
\vdots \\
y_n
\end{bmatrix}
\end{equation}
\begin{equation}
\boldsymbol{\alpha} = (\mathbf{X}^\top\mathbf{X})^{-1}\mathbf{X}^\top\mathbf{Y}
\end{equation}

These features were selected because they were agnostic to the boundary region and could capture the general spatial behavior of the robots and humans in the boundary region. Training data was collected during different days in a large atrium space (15 $\times$ 15 boundary region). During data collection, the robots ran the flocking behavior while a gesturing human moved through the boundary region and the human choreographer selected weight modes on a keyboard. The classification model was trained with several hours of training data. The team used this human-choreographer-annotated supervised dataset to fit a classification problem. During subsequent flocking runs when the predicted weights were used, the model was loaded once during flocking initialization and predictions were queried with a press on the keyboard. As such, the classifier decided which of the seven weight modes to use given the state of the overall system. Any function approximation could have been used in place of the classifier, and our goal was to select a new weight mode similarly to how a human choreographer would have done.

\subsection{Deployment Features}
The team added several features to make the flocking behavior easier to start, operate, and observe. These features included a Bash script to launch the flocking behavior which also included launching auxiliary terminal windows to observe and debug robot states. A second feature was that the robots would remove themselves from the boundary region if their batteries were below a certain threshold so the humans could swap the batteries out. The performance was expected to run for four to eight hours continuously, thus requiring several battery changes during this period. A third feature was enhanced logging, in order to measure how the flocking behavior shifted with different coding changes or human interactants.
\section{Experiment Design}

Two research questions were formulated in order to study the human-robot flocking system and experience:
\begin{itemize}
    \item How do individuals perceive the robot flocking system? How do their perceptions change under different sequences of weight modes?
    \item How do individuals' engage through gesture with the robot flocking system?
\end{itemize}

We designed an experiment to examine these questions. During the experiment, the participant moved throughout the space while the robots were flocking. The participant moved throughout the space three times for five minutes each and completed a survey between each instance. Each of the three flocking instances were separate conditions that corresponded with a different set of weight modes: ``Human Choreographer'', ``Model Prediction'', and ``Control''. In the Human Choreographer condition, an expert choreographer selected the next weight mode from a keyboard while the robots were flocking. The choreographer made these selections based on their viewing of the human participant and the robots' behavior. In the ``Model Prediction'' condition, the model predicted a new weight mode approximately every 30 seconds and updated the robot flocking to perform this predicted weight mode. In the ``Control'' condition, the robots cycled through Cohesion, Separation, and Alignment weight modes in that order for 30 seconds each. Each participant experienced all three weight modes and the conditions were randomized to mitigate ordering effects. 

The experiment lasted approximately 30 minutes and the list of participant tasks was as follows:
\begin{itemize}
    \item \textbf{Intro Survey}: A short survey to confirm the date, their email address, and if they had ever seen the flocking behavior before.
    \item \textbf{Verbal Instructions}: The participants received verbal instructions on how to interact with the robots while they were moving. The full instructions were as follows: \textit{Before you move near the robots I want to share the different ways you can interact with the robots. Feel free to walk among the robots as they move about. If their faces are light blue, they don't see any humans in the space. If they are yellow it means that they see you in the space. They will recognize a few different gestures. The first is you can raise your left hand and the robot will open and close its gripper while its face will illuminate a dark blue. Another gesture you can do is raise your right hand and that will make the robots spin in a circle while its face illuminates green. If they’re too close together while you make this gesture they will pulse green and not spin. The final gesture you can do is put your hands together and the robot will look up to the sky while its face illuminates orange. You can choose to do any or none of these gestures while you interact with the robot. The robots will not only respond to your gestures but also to the locations where you move throughout the space, so keep this in mind while moving around. Do you have any questions?} In addition to these verbal instructions, they received a visual aid to show the list of Human Gestures and the corresponding Robot Responses.
    \item \textbf{Flocking Condition 1}: The participant moved throughout the boundary area for five minutes while the robots flocked under the first randomized condition.
    \item \textbf{Perception Survey 1}: The participant completed a survey to describe their experience during the flocking.
    \item \textbf{Flocking Condition 2}: The participant moved throughout the boundary area for five minutes while the robots flocked under the second randomized condition.
    \item \textbf{Perception Survey 2}: The participant completed a survey that was identical to survey 1 to describe their experience during the flocking.
    \item \textbf{Flocking Condition 3}: The participant moved throughout the boundary area for five minutes while the robots flocked under the third randomized condition.
    \item \textbf{Perception Survey 3}: The participant completed a survey that was identical to surveys 1 and 2 to describe their experience during the flocking.
    \item \textbf{Final Survey}: The participant completed a final survey including demographics and space for additional comments.
\end{itemize}

The Perception Survey consisted of 36 mandatory and 2 optional questions. The first 23 questions were the semantic differential scales for robot perception, also known as the ``Godspeed'' questionnaire \cite{bartneck2008}. The remaining 15 questions were grouped into categories:
\begin{itemize}
    \item \textbf{Gesture Engagement}: There were three questions that inquired about the robot's response to the gestures on a scale of 1 to 5, where 1 was ``Strongly Disagree'' and 5 was ``Strongly Agree'': (I) ``Right Hand Up - Rotate in place. The robot responded correctly to my gesture.'', (II) ``Left Hand Up - Open/close gripper - The robot responded correctly to my gesture.'', and (III) ``Hands Together - gaze up - The robot responded correctly to my gesture.''
    \item \textbf{Robot and Flocking Perception Ratings}: Six questions were ratings questions on a scale of 1 to 5, where 1 was ``Strongly Disagree'' and 5 was ``Strongly Agree'': (IV) ``I would feel comfortable around the robots in this encounter.'', (V) ``I liked the sounds produced by the robots.'', (VI) ``The robots paid attention to me.'', (VII) ``The behavior of the robot was interesting.'', (VIII) ``I enjoyed being around the robots.'', and (IX) ``I would participate in this experience again.''.
    \item \textbf{Robot and Flocking Perception Prompts}: Four questions were open ended writing prompts for the participants to fill in: (1) ``Describe this experience in a few words.'', (2) ``What thoughts came to mind during this experience?'', (3) ``What did you like most about this experience?'', and (4) ``What did you like least about this experience?''
    \item \textbf{Optional Prompts}: Two last questions were open-ended, optional prompts: (1) ``What people do you think would be interested in this experience?'' and (2) ``Any other comments about this experience?''.
\end{itemize}

The team hypothesized that the Gesture Engagement question category would not be different among the experimental conditions, as the head controller and gesture algorithms were not altered between the conditions. Similarly, the team expected the participants' ratings of question ``I liked the sounds produced by the robot (V).'' to be unchanged since all conditions used the same music mode. Question VI, ``The robots paid attention to me.'' was anticipated to be similar across conditions, as it primarily measured the head controller which was constant throughout the experiment. 

As the experimental conditions were meant to alter the robots' behavior through different choices of weight modes, the team hypothesized that the answers to the questions ``I would feel comfortable around the robots in this encounter (IV).'', ``The behavior of the robots was interesting (VII).'', ``I enjoyed being around the robots (VIII).'', and ``I would participate in this experience again (IX).'' would vary with conditions. As the weight modes present across the Control condition led to slower and more close knit robot behavior, the team hypothesized that the Control condition would score the lowest and the Human Choreographer and Model Prediction would be comparable. Extending this hypothesis, the team anticipated that the Human Choreographer and Model Prediction conditions would be rated more highly than the Control condition in terms of Anthropomorphism, Animacy, Likeability, and Perceived Intelligence, but lower in terms of Perceived Safety.

\subsection{Participants}
Participants were recruited by an email sent to a company list. 12 individuals participated in the formal experiment. The participants self-identified their gender and the total report included five men, six women, and one declined to state. Four participants identified their race as White, five were Asian, one was Hispanic or Latino, one was Multiracial, and one was Black or African American. Nine individuals worked at Everyday Robots. 17\% of participants were 18-24, 50\% were 25-34, and 33\% were 35-44. Ten of these participants had previously seen the robots flocking from a higher floor but no participants had moved with the robots while they were flocking.
\section{Results}

\subsection{System Performance}
\begin{figure*}
\centering
\includegraphics[width=\linewidth]{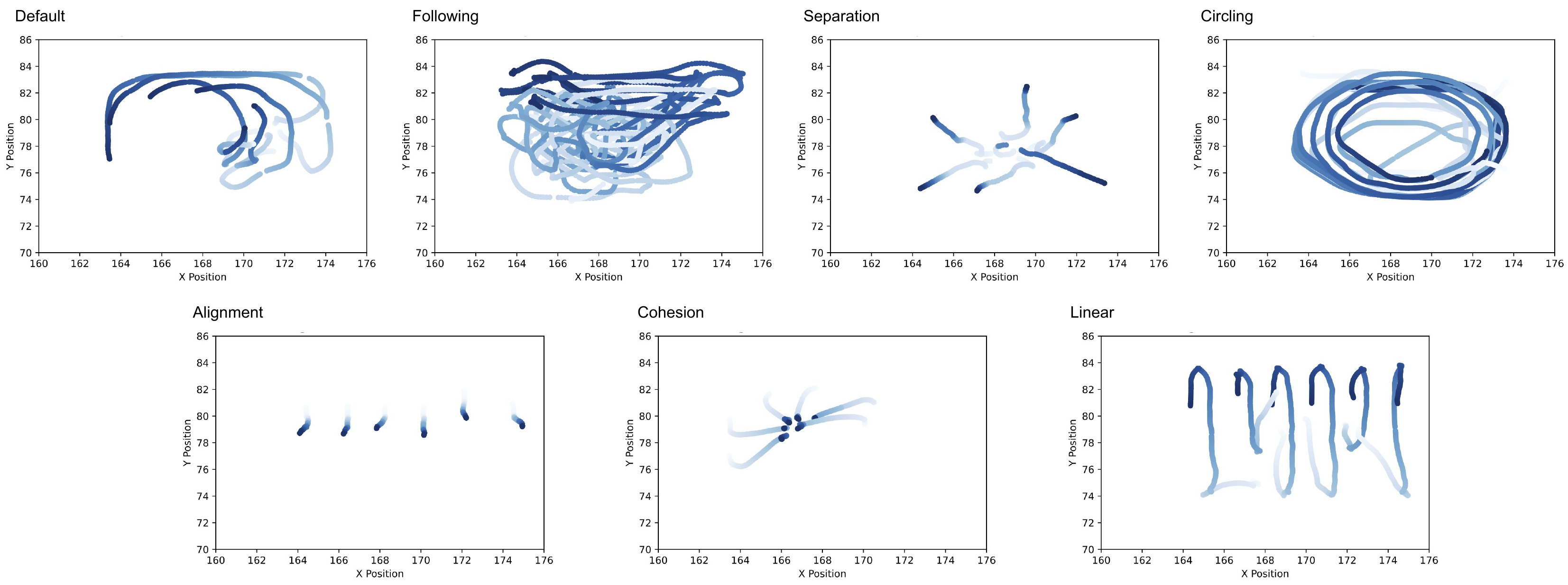}
\caption{Positions of the robots over time during different weight mode settings. From left to right, top row: Default, Following, Separation, and Circling. From left to right, bottom row: Alignment, Cohesion, and Linear.}
\label{weightmode_histogram}
\end{figure*}

The movement of the robots while running different weight modes is shown in Figure \ref{weightmode_histogram}. The robots' movements during different weight modes were visually distinctive, as were the shifts in the robots' behavior when a human entered the scene. The plots show how varied in scale, direction, and speed the robots' movements were under each weight mode. 

\subsection{Survey Results}

\begin{figure*}
\centering
\includegraphics[width=\linewidth]{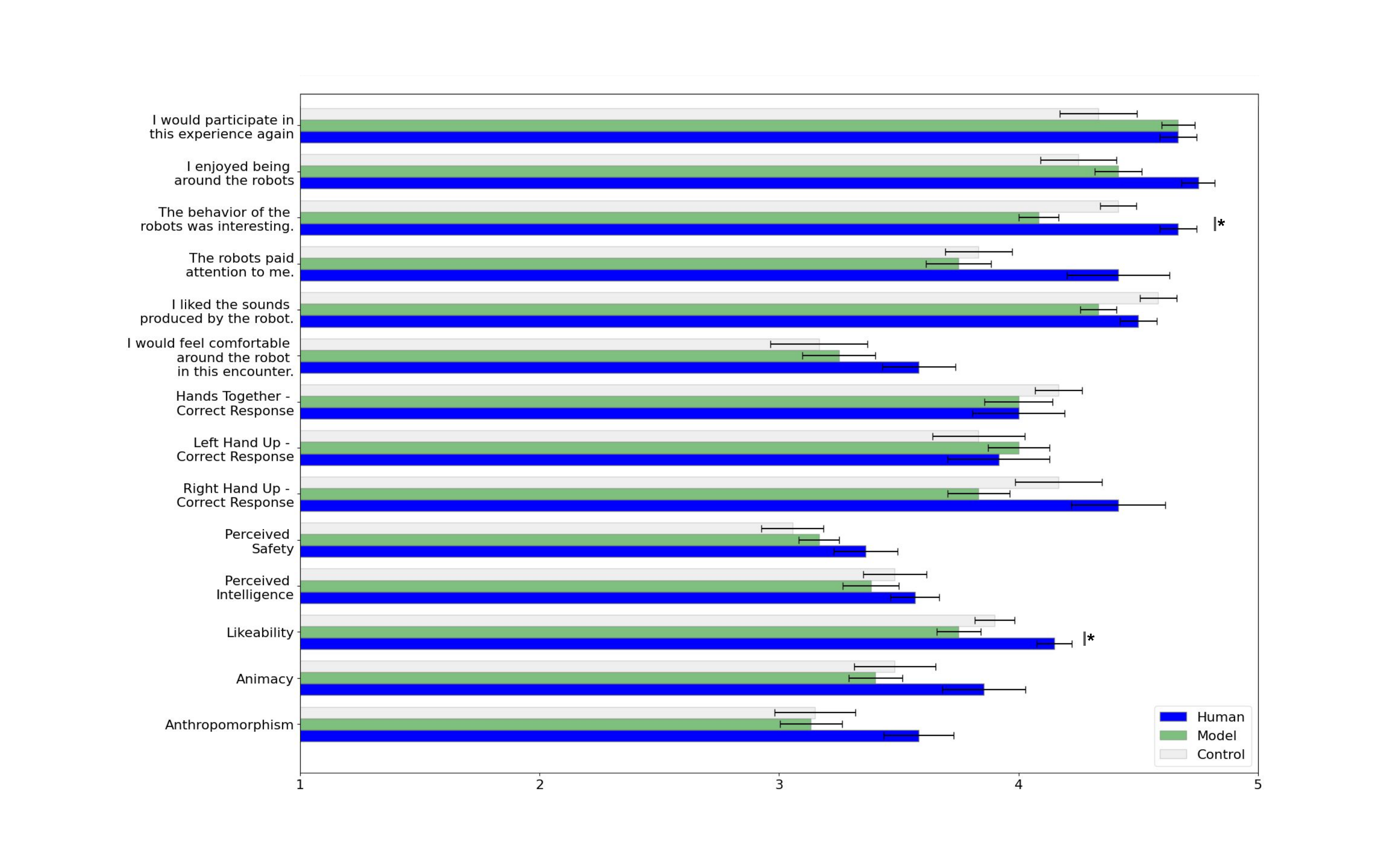}
\caption{Participant ratings for the questions in the Perception Survey. For the Godspeed questions, a repeated measures ANOVA showed statistically significant main effects for the experimental condition on Anthropomorphism and Likeability. The nine remaining quantitative questions were not statistically significant, though question ``The behavior of the robots were interesting'' approached significance. * indicates $p < 0.1$.}
\label{flocking_bartneck_bar}
\end{figure*}

\begin{figure*}
\centering
\includegraphics[width=\linewidth]{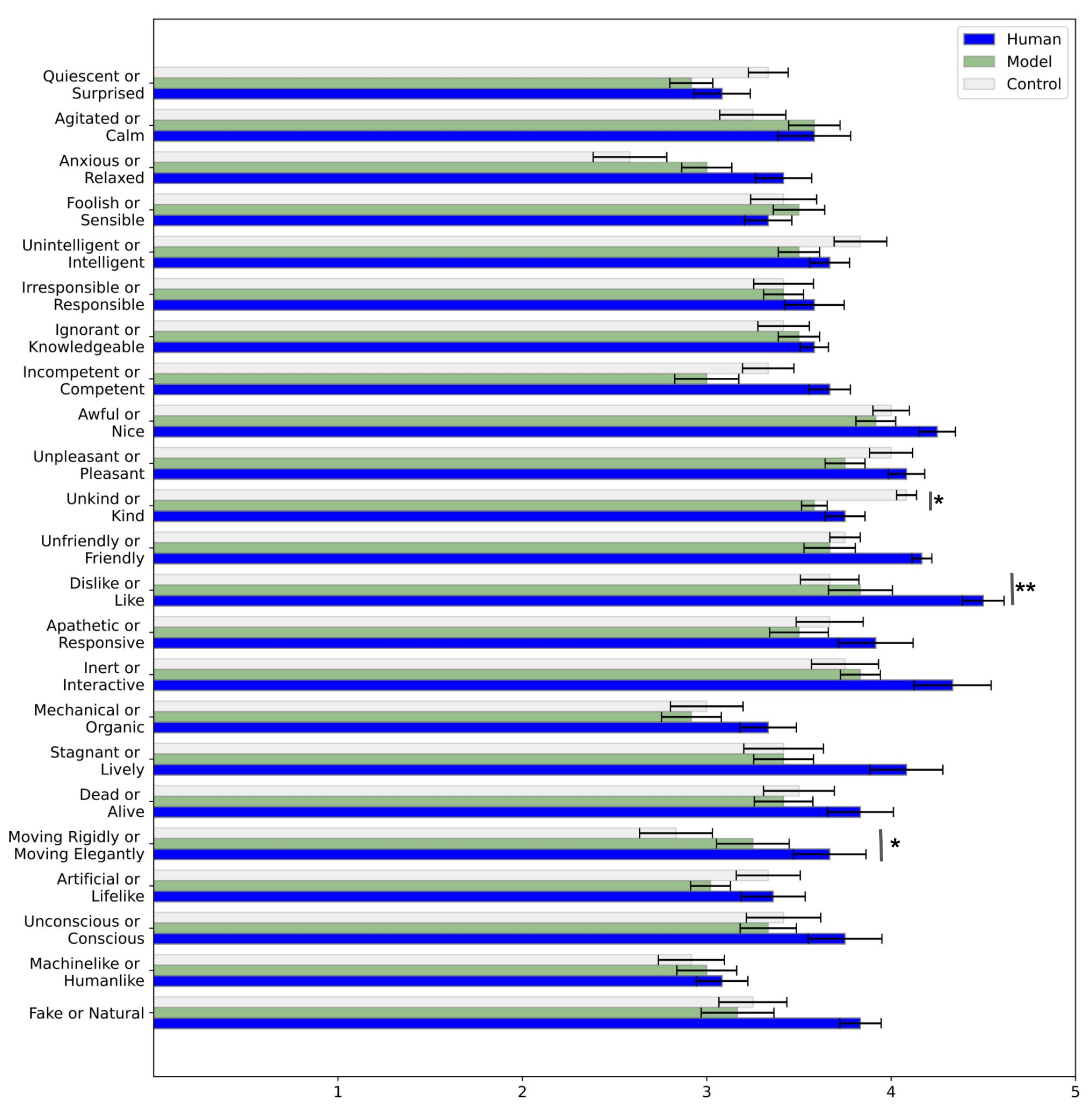}
\caption{Participant ratings for each question from the Godspeed questionnaire. * indicates $p < 0.1$ and ** indicates $p < 0.05$.}
\label{flocking_bartneck_ax}
\end{figure*}

The results from the quantitative questions are shown in Figure \ref{flocking_bartneck_bar} and Figure \ref{flocking_bartneck_ax}. The data for the five Godspeed measures were analyzed with a repeated measures ANOVA. The Model and Human Choreographer conditions were statistically significantly different in Likeability, with (p $< 0.1$). Anthropomorphism, Animacy, Perceived Safety, and Perceived Intelligence were not statistically significant across conditions. These results did not confirm the hypothesis that the Human Choreographer or Trained Model would show a higher score in the Godspeed ratings. Participants were asked to ``Describe the experience in a few words.'' Their descriptions per participant and condition are shown in Table 1. 

A repeated measures ANOVA was used to study the effect of condition on the remaining nine quantitative questions. Questions VIII ``I enjoyed being around the robots.'' and IV ``I would feel comfortable around the robots in this encounter.'' were approaching statistical significance between the Human Choreographer and Model conditions (p $< 0.15$). Question VII ``The behavior of the robots was interesting.'' was statistically significantly different between the Human Choreographer and Model conditions (p $< 0.1$). This confirmed the hypothesis that Gesture Engagement Questions I/II/III, Question V, and Question VI responses would be unchanged between conditions. This did not fully confirm the hypothesis about Questions IV, VII, VIII, and IX changing between conditions, further analysis of this is in the Discussion\ref{discussion} section.

\begin{figure}
\centering
\includegraphics[width=\linewidth]{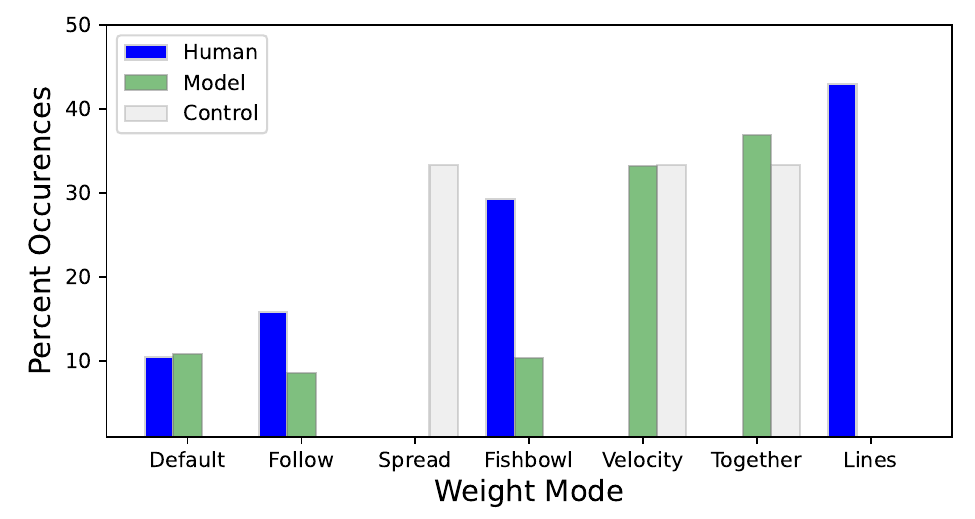}
\caption{Distribution of different weight modes under each condition of the experiment.}
\label{gestures_weights}
\end{figure}

\begin{figure}
\centering
\includegraphics[width=\linewidth]{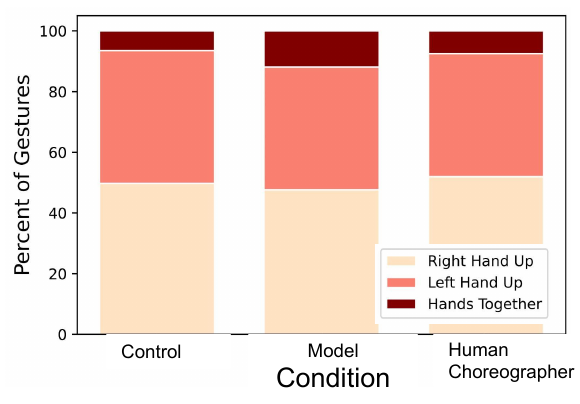}
\caption{The distribution of gestures during each condition.}
\label{gesturedistribution}
\end{figure}

During the experiment, the distribution of the weight modes was different between the Human Choreographer, Model, and Control condition (Figure \ref{gestures_weights}). The Model showed the widest range of weight modes, while the Human Choreographer tended to select from a smaller group. Despite the experimental condition, participants used roughly the same distribution of gestures throughout, as shown in Figure \ref{gesturedistribution}.

\begin{table*}[t]
\caption{Participant responses to \textit{Describe the experience in a few words.}}
\begin{center}
\scalebox{0.85}{
\begin{tabular}{|c|c|c|}
\hline
\textbf{Human} & \textbf{Model} & \textbf{Control}\\
\hline
\hline
Playful, intriguing & Erratic, Miscommunication & Meditative, Herding\\
\hline
Magical & Stubborn bots. & Negotiation\\
\hline
interesting, unique and fun & interesting and exciting & \makecell{It was a fun experience where\\ the interaction with the robot was\\ interesting and exciting but also unique}\\
\hline
chaotic as well & surreal & chaotic\\
\hline
\makecell{The robots were more responsive \\than the previous session. They \\seemed to recognize more gestures} & \makecell{The first attempt was very interesting to\\ learn what the robots perceived and \\that I had to "get their attention" for\\ my movements to be recognized} & \makecell{It seemed as if there were more \\arm/head motions that I did not prompt \\than I noticed previously.}\\
\hline
Endearing but still strange & Less active than the other rounds & Awkward yet fun\\
\hline
Fun, interactive, responsive, surprising & Fun and interactive & Fun, interactive, responsive\\
\hline
\makecell{It was a relaxing and \\meditative experience. i enjoyed \\spending the time together} & \makecell{the robots wouldnt separate from each other.\\ i felt like an other and like they \\didnt want to be near me} & \makecell{i felt ignored. i didn't\\ like this as much}\\
\hline
\makecell{Invigorating, novel, funny} & \makecell{Robots were more clustered \\and kept their distance.} & \makecell{Felt like the robots were\\ slightly more avoidant in \\this condition.}\\
\hline
\makecell{This is super safe, the motions \\are beautiful, i am standing very\\ close to this robot and its planning its \\motions around me } & \makecell{This experience was very \\beautiful, interactive and safe} & \makecell{a very difference use of\\ robots I have seen}\\
\hline
\makecell{Fun, interesting, slow} & \makecell{like being a conductor for \\a group of confused storks} & \makecell{They reminded me of a flock\\ of confused and frightened flamingos}\\
\hline
\makecell{Fun!} & Also great! & \makecell{I was surprised by how \\well the robots responded to\\ my gestures!}\\
\hline

\end{tabular}
}
\end{center}

\end{table*}

\subsection{Qualitative Survey Results and Demonstration Anecdotes}

As this project neared completion, the team tested it in a semi-public space once per week. These testing sessions ran for several weeks and dozens of individuals observed the flocking robots during this time. Passersby included visitors to the building and building inhabitants; visitors included artists, politicians, business consultants, and children while building inhabitants occupied diverse roles like software engineers, program managers, and facilities managers. A number of these individuals sent written feedback to the research team. In addition, the research team took note of any verbal comments from passersby. The majority of these comments were positive and inquisitive. Several passersby remarked upon the sensations they experienced, such as: 
\begin{itemize}
    \item ``I feel like I'm ascending.''
    \item ``Am I in heaven?''
    \item ``The robots are beautiful.''
    \item ``This feels poetic...I'm having a visceral reaction. For me, it's groundbreaking.''
    \item ``They were so cute out there singing and dancing. I have no idea what they were actually meant to be doing but it made me happy.''
\end{itemize}

Other comments were about robots acting similarly to other natural systems, such as: 
\begin{itemize}
    \item ``Little ducklings!''
    \item ``I feel like I'm watching an alien species performing.''
    \item ``Reminded me of watching an aquarium of fish.''
\end{itemize}

Additional comments inquired about the precise functionality of the flocking:
\begin{itemize}
    \item ``I have done projects on teleoperation through movement (with a MagicLeap) before, however I loved the feeling of controlling a fleet through my movements and it felt very different to just controlling one single robot. I liked the overall vibe and atmosphere of it , however I felt a bit worried that the robots might crash into each other or me as they did have some speed (which makes it more vivid I guess?). Anyway, thanks, [this] work is awesome!''
    \item ``Are you [the human] acting as the center average for the robots? 
\end{itemize}
\section{Discussion} \label{discussion}

The results from this experiment indicated that human participants perform the same gestures and have roughly the same perceptions of the robots regardless of the weight mode condition. This may have been because the overall experience of being surrounded by many robots, all moving, playing music, and responding to gestures, was more influential on the participant experience rather than the positions of the robots themselves. While there were small differences in response values for Likeability, Questions IV, VII, and VIII between the Human Choreographer and Model conditions, the change was not convincing enough to conclude that the weight mode selection conditions led to strong changes in participant perception. The lack of variation between the ratings may also have been because the participants' tended to rate all the conditions highly. This is demonstrated by the fact that no single survey question for any single condition averaged below a 2.5. Thus, the participants overall found the robots, and by extension, the system the robots were operating under, to be more Likeable than not, more Animate than not, etc. 

The Model led to the largest variety of weight mode conditions and did not demonstrate the same distribution of weight mode conditions as the Human Choreographer. This may have been because during the experiment, the Human Choreographer may have selected different weight modes than they did during the data collection. The Human Choreographer may have done this because they had access to the real-time, multi-dimensional responses of the human participants. For example, the Human Choreographer could witness that human participants' facial expressions, and selected a different weight mode according to how entertained that individual participant appeared. This type of information was not present during data collection.

\subsection{Conclusion and Future Work}

The primary contributions of this work were a novel group navigation algorithm involving heterogeneous agents, gesture responsiveness for human-robot flocking interaction, a weight mode characterization system, and a learning system for encoding a choreographer's preferences within the overall software system. As this project was designed in part to enthrall and entertain, the emotional, human participant experience was foregrounded as much as the technical contribution. An experiment was designed and conducted to measure how changes in weight modes affected participant reactions and choices of gestures. This work is a demonstration of combining machine learning, controls, and artistic strategies to generate compelling robot behaviors.

In future work, the gesture interactions could be modified to map to different weight modes or learned (rather than choreographed) robot behaviors. The musical samples used in Music Mode could be changed to alter the soundtrack and thus the overall experience. Future experiments could measure changes in participant reactions based on these variations. In addition, more data could be collected to train the weight mode selection agent, and the hand-tuned features could be modified to reflect a richer and broader series of inputs rather than relative positions and geometric representations.

Additionally, a central aim of the project is to present the work in several public forums. This will extend the reach of the project as well as opportunity to study how other individuals interact with these robots in a group human-robot flocking setting.

Copyright \copyright\ \volumeyear\ SAGE Publications Ltd,
1 Oliver's Yard, 55 City Road, London, EC1Y~1SP, UK. All
rights reserved.

\begin{acks}
The authors thank Hans Peter Brondmo and Denise Gamboa for advising and cultivating this project. Thank you to Tom Engbersen, Daniel Lam, and Peter Van Straten for their work on Music Mode. The authors appreciate the facilities team at Rails for their work to prepare the space. Eric Zankiewicz, Tad Koch, and the Everyday Robots support team provided generous robot technical assistance. Will Nail expertly and artistically captured still images and video for this project. Additional thanks to Jeffrey Bingham and Nacho Mellado for research and implementation discussions.
\end{acks}

\section{Declaration of Conflicting Interests}
This work was completed by current and former employees and contractors of Google and Everyday Robots, and researchers at Stanford University. This project was not mandated or led by any other parties at Everyday Robots or Google.

\section{Funding}
The author(s) disclosed receipt of the following financial support for the research, authorship, and/or publication of this article: This work was supported by Google, Inc.

\bibliographystyle{SageH}
\bibliography{bibliography.bib}





\end{document}